\documentclass{article}

\usepackage{arxiv}

\usepackage[utf8]{inputenc} 
\usepackage[T1]{fontenc}    
\usepackage{hyperref}       
\usepackage{url}            
\usepackage{booktabs}       
\usepackage{amsfonts}       
\usepackage{nicefrac}       
\usepackage{microtype}      
\usepackage{lipsum}
\usepackage{graphicx}

\title{Fine-tuning Pre-trained Contextual Embeddings for Citation Content Analysis in Scholarly Publication}

\author{
  Haihua Chen \\
  Department of Information Science\\
  University of North Texas\\
  Denton, TX, USA 76203 \\
  \texttt{haihua.chen@unt.edu} \\
   \And
 Huyen Nguyen \\
  Department of Information Science\\
  University of North Texas\\
  Denton, TX, USA 76203 \\
  \texttt{huyennguyen5@my.unt.edu} \\
}

\begin{document}
\maketitle

\begin{abstract}
 Citation function and citation sentiment are two essential aspects in citation content analysis (CCA), which are useful for influence analysis, the recommendation of scientific publications. However, existing studies are mostly traditional machine learning methods, although deep learning techniques have also been explored, the improvement of the performance seems not significant due to insufficient training data, which brings difficulties to applications. In this paper, we propose to fine-tune pre-trained contextual embeddings ULMFiT, BERT, and XLNet for the task. Experiments on three public datasets show that our strategy outperforms all the baselines in terms of F1 score. For citation function identification, the XLNet model achieves 87.2\%, 86.90\%, and 81.6\% on DFKI, UMICH, and TKDE2019 datasets respectively, while it achieves 91.72\% and 91.56\% on DFKI and UMICH in term of citation sentiment identification. Our method can be used to enhance the influence analysis of scholars and scholarly publications.
\end{abstract}


\section{Introduction}
Citation is a common phenomenon in writing articles, which plays an essential role in academic literature. The process of citing, on one hand, reflects the author's emphasis and interests on others' work, on the other hand, involves existing ideas and methods in his own work. However, traditional citation analysis focuses on quantitative citation measures that they treat all the citations equally. They only show the citation relationship between two articles, but cannot reveal the connotative meaning of citations \cite{hernandez2016survey}. Indeed, in-depth citation content analysis (CCA) using semantic information such as citation function and citation sentiment can make up the deficiencies of traditional citation count-based analysis \cite{zhang2013citation}. Citation function reveals  specific purposes which a citation plays regarding the current paper’s contributions \cite{jurgens2018measuring} while citation sentiment indicates author's attitude and opinion polarity to the cited paper \cite{piao2007mining}. Both of them are very useful for scientific evaluation, citation recommendation, and automatic summarization. However, this idea is challenged for furthering into applications, for it lacks an accurate method to identify citation function and sentiment. Therefore, it remains an open question how to improve the current CCA techniques. 

Existing researches for this task are mainly feature-based traditional machine learning methods. They try to manually identify many different features such as location of the citation sentence, surrounding POS tags, and self-citation \cite{dong2011ensemble}. For example, Meng et al. used word-level features, syntactic features, physical features, and self-citation feature with SVM as the classification algorithm \cite{meng2017automatic}. David et al. divided features into three main categories: pattern-based features, topic-based features, and prototypical argument features \cite{jurgens2018measuring} while Tuarob et al. proposed two heterogeneous sets of features: context and content features. Although their feature sets seemed quite comprehensive, the methods only achieved the best average F1 of 0.749 on citation function identification \cite{8700263}. Also, the performances of all state-of-the-art machine learning based approaches for citation sentiment identification were below 0.9 on accuracy. Recently, deep learning techniques have been applied for useful representation of citations by using word embedding techniques for better citation function and sentiment identification \cite{lauscher2017investigating,yousif2019multi,perier2019preliminary}; however, they have not shown many significant improvements due to insufficient training data. Besides, most of the existing approaches focus on either citation function identification or citation sentiment identification, so it brings many difficulties to applications.     

To bridge the gap, we propose to fine-tuning pre-trained contextual embeddings ULMFiT \cite{howard2018universal}, BERT \cite{devlin2018bert}, and XLNet \cite{yang2019xlnet} for citation function and sentiment identification since the language models are trained in the general domains which have different data distributions from the target domains and tasks. Moreover, fine-tuning pre-trained language models have been proved to be effective to enhance the performance on text classification \cite{sun2019fine}. We design experiments on three public dataset to evaluate the proposed strategy. The contributions of our paper are as follows:

\begin{itemize}
\item We propose a new solution to enhance citation function and sentiment identification by fine-tuning the pre-trained contextual embeddings.
\item We achieve state-of-the-art results on three public available dataset, demonstrating the robust of our proposed method.  
\end{itemize}

\section{Related Work}
\label{sec:headings}

Since CCA have been proven to be a promising strategy to replace existing quantitative measurements such as h-index in measuring the influence of scholars and scholarly publication, efforts have been made to automatically analyze citations by using both supervised learning \cite{meng2017automatic, jia2018citation}, and unsupervised learning method \cite{ding2014content}. Most of these models relied on manually engineered features. Most feature-based classification models discovered the following features: Cue phases built by part-of-speech-based tags \cite{Teufel:2006:ACC:1610075.1610091, meng2017automatic, jia2018citation} and n-grams as semantic features  \cite{meng2017automatic, jia2018citation}, dependency relation as syntactic features \cite{meng2017automatic, jia2018citation}, Boolean \cite{dong2011ensemble, meng2017automatic}, and other physical features such as location and frequency. A variety of traditional machine learning classifiers were utilized to automatically annotate citations such as IBk machine learning \cite{Teufel:2006:ACC:1610075.1610091}, support vector machines (SVMs) \cite{athar2012context, meng2017automatic, jia2018citation}, Maximum entropy \cite{jia2018citation}, etc. SVM classifier performed the best in some experiments \cite{jia2018citation, meng2017automatic}. \cite{meng2017automatic} worked on the same dataset (DFKI) and classification scheme as \cite{dong2011ensemble}, but  applying the powerful classifier SVM and discovering more useful features, they got an improvement of 20\% higher than the study of \cite{dong2011ensemble}. Class imbalance had been a problem of automatic citation classification; for example, background functions and neutral sentiments dominated datasets rather than negative sentiments or any other citation functions \cite{dong2011ensemble}.

Nevertheless, the above feature-based citation classification models were quite time-consuming to engineer numerous features and limited in small training data \cite{yousif2019multi}. Furthermore, engineered features were too specific in a domain to be able to apply in others \cite{lauscher2017investigating}. Deep learning, a branch of the neural network, has recently been applied to tackle the above problems of citation classification \cite{lauscher2017investigating, yousif2019multi,cohan2019structural}. The deep learning method uses multiple layers to extract and transform features hierarchically. Instead of manually engineering features as the previous citation classification models did, this method utilized word-embedding technique to transfer words into vectors \cite{zhang2018deep}. Glove word embedding was applied as a pre-training technique of non-contextualized embedding \cite{yousif2019multi, cohan2019structural, lauscher2017investigating}. Pre-trained language models such as ELMo \cite{cohan2019structural} and BERT \cite{zhao2019context} have been used effectively for contextualized embedding. To the best of our knowledge, the model of Yousif et al. have achieved the best result so far in both dimensions of citation classifications, but it worked better in citation sentiment (87.91\% on F-score) than in citation function (84.62\%) \cite{yousif2019multi}.

Although existing research has made certain progress in citation function and sentiment identification, they completely depend on different datasets and different classification schemes of citation function and sentiment. Given the value of citation function and sentiment in scientific evaluation, it is important to develop an universal and effective approach in this task. Therefore, our proposal can not only model contextual information but also fill the gap of insufficient labeled data, which we argue can help improve the performance of citation function and sentiment identification. 

\section{Methodology}
\label{sec:others}

Our proposed approach is based on pre-trained contextual embeddings such as ULMFiT \cite{howard2018universal}, BERT \cite{devlin2018bert}, and XLNet \cite{yang2019xlnet}, which have shown to be useful in text classification when labelled data is insufficient, since they can utilize a large amount of unlabelled data. When we fine-tune these embeddings for a target task, several components should be taken into consideration: Whether further pre-training is needed for the task? Which layer is better for the task since different layers  capture different levels of semantic and syntactic information? How to choose a better optimization algorithm and learning rate? \cite{sun2019fine}

\begin{figure}
    \centering
    \includegraphics[width=0.8\textwidth]{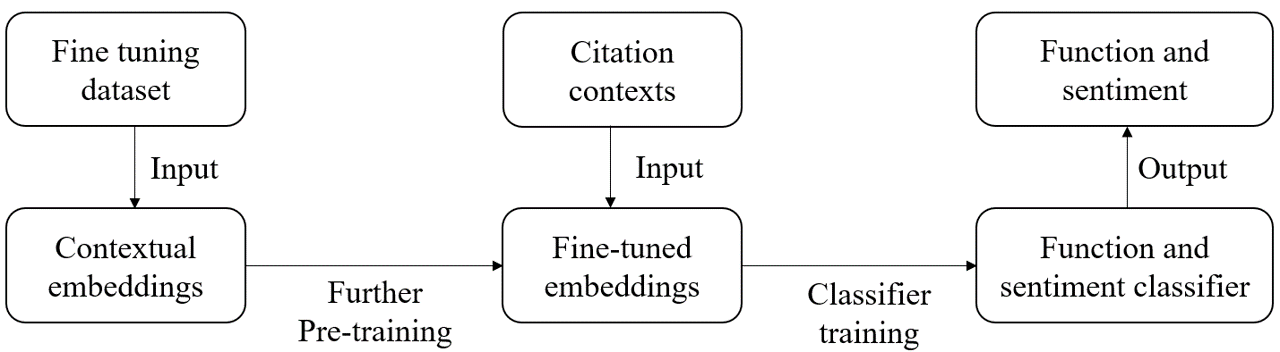}
    \caption{Overview of fine-tuning contextual embeddings for citation function and sentiment identification}
    \label{fig:fig1}
\end{figure}

The workflow of fine-tuning these pre-trained contextual embeddings for citation function and sentiment identification is shown in figure \ref{fig:fig1}. In our strategy, we first further fine-tune the pre-trained embeddings to a new dataset in such a manner that it does not forget what it previously learned. What's more, recent research on transfer learning using pre-trained contextual embeddings has demonstrated that freezing the majority of the weights when fine-tuning results in only minimal accuracy declines. Therefore, we only add a single linear layer on top of the pre-trained model for classification, which performs as sentence classifier. As we feed input data, the entire pre-trained model and the additional untrained classification layer is trained on our citation function and sentiment identification task. 

\section{The Experiments and Evaluation}
\label{sec:experiment}
To test the effectiveness and robust of the proposed deep transfer learning models on the citation function and sentiment identification task, we conduct experiments and evaluation on four public datasets. In this section, we present the performance of the proposed method on four public datasets and compare with the state-of-the-art approaches.

\subsection{Datasets}

We utilize the following three datasets in our experiments for citation function and sentiment identification. Among them, DFKI \cite{dong2011ensemble} contains 1768 instances and UMICH \cite{jha2017nlp} contains 3568 citation context examples labelled with citation function and sentiment, while TKDE \cite{8700263} 8,796 citation contexts which are only labelled with citation functions. A descriptive statistics of the three datasets is presented in table 1. 



\begin{table}[h]
\caption{Descriptive statistics of the corpora}
\label{table:corpora}
\begin{center}
\begin{tabular}{lp{4cm}p{4cm}}
\hline
\textbf{Dataset}&\textbf{Citation sentiment}&\textbf{Citation function}\\
\hline
DFKI\cite{dong2011ensemble} & Positive: 10.75\% \newline Negative: 3.22\% \newline Neutral: 86.03\% & Idea: 7.18\% \newline Basis: 23.81\% \newline GRelated: 42.48\% \newline SRelated: 20.81\% \newline MRelated: 1.75\% \newline Compare: 3.97\%\\
\hline
UMICH\cite{jha2017nlp} & Positive: 32.6\% \newline Negative: 12.4\% \newline Neutral: 55.0\% & Criticizing: 16.3\% \newline Comparison: 8.1\% \newline Use: 18.0\% \newline Substantiating: 8.0\% \newline Basis: 5.3\% \newline Neutral: 44.3\%\\
\hline
TKDE\cite{8700263} &  & Use: 8.55\% \newline Extend: 4.30\% \newline Mention: 65.37\% \newline Notalgo: 21.78\%\\
\hline
\end{tabular}
\end{center}
\end{table}

\subsection{Experiments Setup}
Our proposed fine-tuning approach is based on ULMFiT, BERT, and XLNet, respectively. For ULMFiT, we use AWD-LSTM model with an embedding size of 400, 3 layers and 1150 hidden activations per layer \cite{howard2018universal}. For BERT, we use the BERT-base model with a hidden size of 768, 12 layers and 12 self-attention heads \cite{sun2019fine}. For XLNet, we also use XLNet-base model with 12 layers, 768 hiddens, 12 heads. 

We fine-tune the three models on 1 NVIDIA Quadro P5000 GPU and set the batch size to 32, max squence length of 128 and learning rate of 2e-5 to ensure that the GPU memory is fully utilized. The dropout probability is always kept at 0.1. We use Adam with $\beta_1$ = 0.9 and $\beta_2$ = 0.999. We empirically set the max number of the epoch to 16 and save the best model on the validation set for testing. We conduct 10-fold cross validation to avoid over-fitting.

\subsection{Baselines and Results}

We compare the proposed fine tuning strategy with the following baselines: \cite{lauscher2017investigating}, \cite{jha2017nlp}, \cite{yousif2019multi}, and \cite{8700263} are for citation function identification, while \cite{lauscher2017investigating}, \cite{DBLP:journals/corr/GhoshD017}, \cite{jha2017nlp}, \cite{yousif2019multi}, and \cite{ikram2019aspect}, are for citation sentiment identification. Among which the first seven baselines are feature based machine learning methods while the rest two are deep learning based methods. As always, we use F1 as an evaluation indicator.

\begin{table}[h]
\caption{The results of citation function identification on different corpora}
\label{table:result}
\begin{center}
\begin{tabular}{p{2.8cm}|p{2cm}|p{2cm}|p{2cm}}
\hline
Model&DFKI&UMICH&TKDE2019 \\
\hline
Lauscher et al. \cite{lauscher2017investigating}& 74.30\% & {--} & -- \\
\hline
Rahul et al. \cite{jha2017nlp}& {--} & 64.96\% & -- \\
\hline
Yousif et al. \cite{yousif2019multi} &84.62\%&83.08\%& -- \\
\hline
Tuarob et al. \cite{8700263} & {--} & -- & 74.90\% \\
\hline
ULMFit fine tuning & 85.34\% & 84.39\% & 78.62\% \\
\hline
BERT fine tuning & 85.88\% & 85.92\%& 80.24\% \\ 
\hline
XLNet fine tuning & 87.2\%& 86.90\% & 81.6\% \\ 
\hline
\end{tabular}
\end{center}
\end{table} 

The experiment results on citation function identification are shown in table 2. Overall, the our proposed strategies outperform feature based machine learning and deep learning methods, as well as pre-trained contextual embeddings without fine-tuning, indicating the effectiveness of fine-tuning pre-trained contextual embeddings in this task. Since previous study already proved the usefulness of pre-trained contextual embeddings \cite{cohan2019structural, zhao2019context} in citation function identification, our experiments further demonstrate the necessity of fine tune pre-trained embeddings trained in the general domain to adapt to the downstream classification task. 

When analyzing the results in depth we notice that fine-tuning XLNet-base model performs better than fine tuning the other two embeddings, suggesting the advantages of XLNet over BERT and ULMFiT given the same training conditions. The reason can be referred from the statement in the XLNet paper \cite{yang2019xlnet}, XLNet integrates ideas from Transformer-XL into pretraining, which can overcome the limitations of BERT thanks to its autoregressive formulation. Meanwhile, compared to BERT, ULMFiT is biased towards negative for shorter sentences, such as the citation contexts in the citation function identification task. 

\begin{table}[h]
\caption{The results of citation sentiment identification on different corpora}
\label{table:result}
\begin{center}
\begin{tabular}{p{3cm}|p{3cm}|p{3cm}}
\hline
Model&DFKI&UMICH \\
\hline
Lauscher et al. \cite{lauscher2017investigating}& 78.80\% & --  \\
\hline
Rahul et al. \cite{jha2017nlp}& {--} & 78.50\%  \\
\hline
Souvick et al. \cite{DBLP:journals/corr/GhoshD017} &83.38\%& {--} \\
\hline
Yousif et al. \cite{yousif2019multi}  & 87.91\% & 85.68\% \\
\hline
Ikram et al. \cite{ikram2019aspect} & 75.00\% &{--} \\
\hline
ULMFit fine tuning& 88.40\% & 86.32\% \\
\hline
BERT fine tuning & 90.87\% & 89.90\%\\
\hline
XLNet fine tuning& 91.72\%& 91.56\%\\
\hline
\end{tabular}
\end{center}
\end{table} 

The experiment results on citation sentiment identification are shown in table 3. We can draw the same conclusion as citation function identification: XLNet fine tuning achieve the best performance on citation sentiment identification. The experiments results not only indicate pre-trained embeddings are useful for citation content analysis, but also demonstrate that XLNet fine tuning could be one of the most effective models to develop the automatic citation content analysis tools.  

\section{Summary and future work}

In this paper, we propose a deep learning method which fine-tune pre-trained contextual embeddings for CCA. For the word embeddings, we use ULMFiT,BERT, and XLNet, the unlabelled text is applied to further fine-tune the embeddings, then classifiers are trained using the labelled data. Our experiment on three public datasets show that XLNet outperforms the baseline methods and other word embeddings, by achieving 87.2\%, 86.90\%, 81.6\% on DFKI, UMICH, TKDE2019 datasets for citation function identification, and 91.72\%, 91.56\% on DFKI, UMICH for citation sentiment identification respectively. 

In the future, it would be interesting to investigate how the fine tuning datasets such as filed and amount could affect the performance of automatic CCA. After identifying the best model, we can consider to collect large-scale of scholars and scholarly publications for further evaluation, and finally implement the automatic CCA applications.

\bibliographystyle{unsrt}  


\begin{thebibliography}{1}

\bibitem{athar2012context}
Awais Athar and Simone Teufel. 2012.
\newblock Context-enhanced citation sentiment detection. 
\newblock In {\em Proceedings of the 2012 conference of the North American chapter of the Association for Computational Linguistics: Human language technologies. } Association for Computational Linguistics,
597–601.


\bibitem{cohan2019structural}
Arman Cohan, Waleed Ammar, Madeleine van Zuylen, and Field Cady. 2019.
\newblock Structural Scaffolds for Citation Intent Classification in Scientific Publications.
\newblock {\em arXiv preprint arXiv:1904.01608}, 2019.


\bibitem{devlin2018bert}
Jacob Devlin, Ming-Wei Chang, Kenton Lee, and Kristina Toutanova. 2018.
\newblock Bert: Pre-training of deep bidirectional transformers for language understanding.
\newblock {\em arXiv preprint arXiv:1810.04805}, 2018.


\bibitem{ding2014content}
Ying Ding, Guo Zhang, Tamy Chambers, Min Song, Xiaolong Wang,
and Chengxiang Zhai. 2014.
\newblock Content-based citation analysis: The
next generation of citation analysis.
\newblock {\em Journal of the Association for
Information Science and Technology}, 65, 9 (2014), 1820–1833.


\bibitem{dong2011ensemble}
Cailing Dong and Ulrich Schäfer. 2011.
\newblock Ensemble-style self-training on
citation classification. 
\newblock In {\em Proceedings of 5th international joint conference on natural language processing.} 623–631.


\bibitem{DBLP:journals/corr/GhoshD017}
Souvick Ghosh, Dipankar Das, and Tanmoy Chakraborty. 2017. 
\newblock Determining sentiment in citation text and analyzing its impact on the proposed ranking index. 
\newblock {\em arXiv preprint arXiv:1707.01425}, 2017.


\bibitem{hernandez2016survey}
Myriam Hernández-Alvarez and José M Gomez. 2016.  
\newblock Survey about citation context analysis: Tasks, techniques, and resources. 
\newblock {\em Natural
Language Engineering}, 22, 3 (2016), 327–349.


\bibitem{howard2018universal}
Jeremy Howard and Sebastian Ruder. 2018.   
\newblock Universal language model
fine-tuning for text classification. 
\newblock {\em arXiv preprint arXiv:1801.06146}, 2018.


\bibitem{ikram2019aspect}
Muhammad Touseef Ikram and Muhammad Tanvir Afzal. 2019.    
\newblock Aspect based citation sentiment analysis using linguistic patterns for better comprehension of scientific knowledge. 
\newblock {\em Scientometrics}, 119, 1 (2019), 73–95.


\bibitem{jha2017nlp}
Rahul Jha, Amjad-Abu Jbara, Vahed Qazvinian, and Dragomir R Radev.
2017.    
\newblock NLP-driven citation analysis for scientometrics. 
\newblock {\em Natural Language Engineering}, 23, 1 (2017), 93–130.


\bibitem{jia2018citation}
Meng Jia. 2018.   
\newblock Citation function and polarity classification in biomedical papers.  
\newblock {}2018.


\bibitem{jurgens2018measuring}
David Jurgens, Srijan Kumar, Raine Hoover, Dan McFarland, and Dan Jurafsky. 2018.    
\newblock Measuring the evolution of a scientific field through
citation frames. 
\newblock {\em Transactions of the Association for Computational
Linguistics}, 6 (2018), 391–406.


\bibitem{lauscher2017investigating}
Anne Lauscher, Goran Glavaš, Simone Paolo Ponzetto, and Kai Eckert. 2017.    
\newblock Investigating convolutional networks and domain-specific embeddings for semantic classification of citations.  
\newblock In {\em Proceedings of
the 6th International Workshop on Mining Scientific Publications.}, ACM, 24–28.


\bibitem{meng2017automatic}
Rui Meng, Wei Lu, Yu Chi, and Shuguang Han. 2017.    
\newblock Automatic classification of citation function by new linguistic features.   
\newblock In {\em iConference
2017 Proceedings.} 2017.


\bibitem{perier2019preliminary}
Julien Perier-Camby, Marc Bertin, Iana Atanassova, and Frédéric
Armetta. 2019.     
\newblock A preliminary study to compare deep learning with rulebased approaches for citation classification.   
\newblock In {\em 8th International Workshop on Bibliometric-enhanced Information Retrieval (BIR) co-located
with the 41st European Conference on Information Retrieval (ECIR 2019)}, Vol. 2345. 125–131.


\bibitem{piao2007mining}
Scott Piao, Sophia Ananiadou, Yoshimasa Tsuruoka, Yutaka Sasaki, and John McNaught. 2007.     
\newblock A preliminary study to compare deep learning with rulebased approaches for citation classification.   
\newblock In {\em 8th International Workshop on Bibliometric-enhanced Information Retrieval (BIR) co-located
with the 41st European Conference on Information Retrieval (ECIR 2019)}, Vol. 2345. 125–131.


\bibitem{sun2019fine}
Chi Sun, Xipeng Qiu, Yige Xu, and Xuanjing Huang. 2019.    
\newblock How to FineTune BERT for Text Classification?   
\newblock {\em arXiv preprint arXiv:1905.05583}, 2019.


\bibitem{Teufel:2006:ACC:1610075.1610091}
Teufel, Advaith Siddharthan, and Dan Tidhar. 2006.
\newblock Automatic Classification of Citation Function.   
\newblock In {\em Proceedings of the 2006 Conference on Empirical Methods in Natural Language Processing (EMNLP
’06).} Association for Computational Linguistics, Stroudsburg, PA, USA, 103–110. http://dl.acm.org/citation.cfm?id=1610075.1610091.


\bibitem{8700263}
S. Tuarob, S. W. Kang, P. Wettayakorn, C. Pornprasit, T. Sachati, S. Hassan, and P. Haddawy. 2019. 
\newblock Automatic Classification of Algorithm Citation Functions in Scientific Literature.   
\newblock {\em IEEE Transactions on Knowledge and Data Engineering (2019)}, 1–1. https://doi.org/10.1109/TKDE.
2019.2913376.


\bibitem{yang2019xlnet}
Zhilin Yang, Zihang Dai, Yiming Yang, Jaime Carbonell, Ruslan
Salakhutdinov, and Quoc V Le. 2019. 
\newblock XLNet: Generalized Autoregressive Pretraining for Language Understanding.   
\newblock {\em arXiv preprint
arXiv:1906.08237}, 2019.


\bibitem{yousif2019multi}
Abdallah Yousif, Zhendong Niu, James Chambua, and Zahid Younas
Khan. 2019. 
\newblock Multi-task learning model based on recurrent convolutional neural networks for citation sentiment and purpose classification.    
\newblock {\em Neurocomputing}, 335 (2019), 195–205.


\bibitem{zhang2013citation}
Guo Zhang, Ying Ding, and Staša Milojević. 2013.  
\newblock Citation content analysis (CCA): A framework for syntactic and semantic analysis of citation content.     
\newblock {\em Journal of the American Society for Information Science and Technology}, 64, 7 (2013), 1490–1503.


\bibitem{zhang2018deep}
Lei Zhang, Shuai Wang, and Bing Liu. 2018.  
\newblock Deep learning for sentiment analysis: A survey.      
\newblock {\em Wiley Interdisciplinary Reviews: Data Mining
and Knowledge Discovery}, 8, 4 (2018), e1253.


\bibitem{zhao2019context}
He Zhao, Zhunchen Luo, Chong Feng, Anqing Zheng, and Xiaopeng
Liu. 2019. 
\newblock A Context-based Framework for Modeling the Role and
Function of On-line Resource Citations in Scientific Literature.
\newblock In {\em Proceedings of the 2019 Conference on Empirical Methods in Natural
Language Processing and the 9th International Joint Conference on Natural Language Processing (EMNLP-IJCNLP). } 5209–5218.

\end{thebibliography}

\end{document}